\documentclass[11pt, a4paper]{article}
\usepackage[T1]{fontenc}
\usepackage{xcolor} %added by noah for color font

\usepackage{graphicx}
\usepackage{amssymb}
\usepackage{amsmath}
\usepackage{booktabs}
\usepackage{tabularx}
\usepackage{makecell}   % for multi-line cells
\usepackage{array}
\usepackage{multirow}
\usepackage{setspace}
\usepackage[hidelinks]{hyperref}

\usepackage{todonotes}
\usepackage{authblk}

\begin{document}
\title{Fine-Tuning A Large Language Model for Systematic Review Screening}
\author[1]{Kweku Yamoah}
\author[1]{Noah Schroeder\thanks{Corresponding author: schroedern@ufl.edu}}
\author[1]{Emmanuel Dorley}
\author[1]{Neha Rani}
\author[1]{Caleb Schutz}

\affil[1]{University of Florida, Gainesville, FL, 32611}

%\institute{University, City State Zipcode, USA 
%\email{\{email 1, email 2\}@email.edu}}
%
\maketitle              % typeset the header of the contribution
\begin{abstract}
Systematic reviews traditionally have taken considerable amounts of human time and energy to complete, in part due to the extensive number of titles and abstracts that must be reviewed for potential inclusion. Recently, researchers have begun to explore how to use large language models (LLMs) to make this process more efficient. However, research to date has shown inconsistent results. We posit this is because prompting alone may not provide sufficient context for the model(s) to perform well. In this study, we fine-tune a small 1.2 billion parameter open-weight LLM specifically for study screening in the context of a systematic review in which humans rated more than 8500 titles and abstracts for potential inclusion. Our results showed strong performance improvements from the fine-tuned model, with the weighted F1 score improving 80.79\% compared to the base model. When run on the full dataset of 8,277 studies, the fine-tuned model had 86.40\% agreement with the human coder, a 91.18\% true positive rate, a 86.38\% true negative rate, and perfect agreement across multiple inference runs. Taken together, our results show that there is promise for fine-tuning LLMs for title and abstract screening in large-scale systematic reviews. 

%\keywords{Large Language Model  \and Fine-tuning \and Systematic Review.}
\end{abstract}
\section{Introduction}
Systematic reviews are a critical method in educational technology research, offering researchers synthesized findings across an entire field of study. However, systematic reviews are inherently time-consuming \cite{borah2017analysis,cao2025automation}, with one study estimating that reviews take on average 67 weeks to complete \cite{borah2017analysis}. Much of the time required for systematic reviews comes from the first stage of study screening, when titles and abstracts are reviewed. Large-scale, comprehensive syntheses may have thousands of titles and abstracts that need to be reviewed for potential inclusion. For example, the authors are currently working on a systematic review where they screened more than 8,500 titles and abstracts. Such a process is time and resource intensive. 

Researchers have been exploring how large language models (LLMs) can be used to help alleviate the burden of title and abstract screening. One large scale study found that LLMs can potentially reduce the human workload during this phase of systematic reviews by up to 93\% \cite{delgado2025transforming}. However, LLMs are heavily context dependent and can be strongly influenced by the wording of the prompt and other parameter settings, and not all models will perform identically even with the same parameter settings or prompt \cite{cao2025development,delgado2025transforming}. Consequently, while promising, the current state of the art is not to fully rely on LLMs during title and abstract screening \cite{Flemyng25,li2024evaluating}.

While the context-dependence of of LLMs has often been viewed as as a quality that prohibits their widespread adoption in the systematic review process, we argue that it could instead be beneficial to embrace their context-dependence. In short, we hypothesized that fine-tuning a small LLM based on human-coded data from \textit{one specific review} could lead to higher performing and more consistent LLM inferences for that review. Fine-tuning a small LLM is an easily approachable task using publicly available code, and small LLMs can be run and trained on commonly available consumer hardware \cite{zhang2024,belcak2025}. Moreover, the process of title and abstract screening during systematic reviews creates the foundation for well-formatted datasets for training models. The purpose of this study was to explore how fine-tuning a small LLM can impact its performance during LLM-based screening of titles and abstracts for an individual systematic review. 

\subsection{The Present Study}
We explored if fine-tuning a small LLM for a specific systematic review could achieve strong and reliable results. If so, it could greatly reduce the amount of time required to conduct systematic reviews. Specifically, we sought to answer the following research questions: 
\begin{enumerate}
    \item What is the model's baseline performance at title and abstract screening compared to human ratings?
    \item How does a fine-tuned, small LLM perform on a small validation dataset compared to human ratings?
    \item How does a fine-tuned, small LLM perform across a large, full dataset compared to a human rater?
\end{enumerate}
\section{Methods}
\subsection{The Dataset and Human Coding}
The dataset consisted of the titles and abstracts from 8,694 studies that were located as part of a systematic review on the influence of generative artificial intelligence technologies in undergraduate computer science education. Specifically, the dataset consists of titles, abstracts, and human-coded inclusion or exclusion decisions (coded as `1' or `0', respectively). 
The entire dataset was coded by humans for inclusion or exclusion in the systematic review. Three individuals participated in the coding process. In order to calculate inter-rater agreement, 273 titles and abstracts were coded by all three coders. The inter-rater agreement exceeded 99\%.

\subsection{The LLM and Fine-Tuning}
We fine-tuned Liquid AI's \texttt{LFM2.5-1.2B-Instruct} in bfloat16 precision~\cite{huggingfaceLiquidAILFM2512BInstructHugging}. The model was selected because its compact size (1.2B parameters) enables local training and evaluation on single-GPU hardware while retaining strong instruction-following capability.

\subsubsection{Fine-Tuning Approach}
We performed supervised fine-tuning (SFT) using \emph{full-parameter fine-tuning}, i.e., updating \emph{all} model weights rather than training parameter-efficient adapters. While parameter-efficient fine-tuning (PEFT) methods such as LoRA can be highly effective, prior comparative studies show that PEFT and full fine-tuning exhibit meaningful trade-offs in both performance and computational cost, and PEFT can underperform full fine-tuning depending on task difficulty, data regime, and language setting~\cite{razuvayevskaya2024peftfft,thakkar2024pefttradeoffs,lialin2024peftsurvey}. Because our setting involves a single target task and a relatively small base model, full fine-tuning is feasible and avoids adapter-induced capacity constraints, allowing the model to realize the highest adaptation ceiling for the task.

\subsubsection{Training Stack and Efficiency (Unsloth)}
We implemented SFT using Unsloth integrated with the Hugging Face/TRL training stack. Unsloth provides optimized kernels (e.g., Triton-based implementations and other training-time optimizations) that reduce memory use and accelerate fine-tuning while remaining compatible with standard Hugging Face tooling~\cite{hfblogUnslothTRL2024,hfTransformersUnslothIntegration,trlUnslothIntegration,unslothDocs}. %In particular, official documentation and integration guides report \(\sim\!2\times\) faster training and substantial VRAM reductions (often 60--80\% depending on model and configuration), which is especially valuable for single-GPU experimentation and reproducible local runs~\cite{unslothDocs,trlUnslothIntegration,hfblogUnslothTRL2024}.

\subsubsection{Training Configuration}
We configured training using the hyperparameters in Table~\ref{tab:training_config}. %We additionally enabled response-only loss masking (computing loss only on assistant tokens) to align optimization with the intended binary output format.

\paragraph{Full fine-tuning configuration.}
Full fine-tuning was enabled in Unsloth \\(\texttt{full\_finetuning=True}), resulting in all model parameters being trainable (100\% of weights updated). We used bfloat16 mixed precision for efficiency and stability on modern accelerators, and an 8-bit AdamW optimizer to reduce optimizer-state memory footprint.

    \begin{table}[t]
    \centering
    \caption{\textbf{Fine-tuning configuration for \texttt{LFM2.5-1.2B-Instruct}.} We report the training regime, hyperparameters, and implementation choices used in our experiments.}
    \label{tab:training_config}
    \tiny
    \singlespacing
    \setlength{\tabcolsep}{4pt}
    \renewcommand{\arraystretch}{1.08}
    \begin{tabularx}{\linewidth}{@{}>{\raggedright\arraybackslash}p{3.2cm} >{\raggedright\arraybackslash}p{2.4cm} X@{}}
    \toprule
    \textbf{Parameter} & \textbf{Value} & \textbf{Rationale} \\
    \midrule
    \multicolumn{3}{@{}l}{\textit{Fine-tuning regime}} \\
    Adaptation method & Full fine-tuning & Update all weights (no adapters) to maximize task-specific capacity~\cite{razuvayevskaya2024peftfft,thakkar2024pefttradeoffs}. \\
    \addlinespace[2pt]
    \hline
    
    \multicolumn{3}{@{}l}{\textit{Optimization}} \\
    \addlinespace[1pt]
    Optimizer & AdamW (8-bit) & Reduces optimizer-state memory while preserving AdamW behavior~\cite{dettmers20228bitoptimizers}. \\
    Learning rate & $2\times 10^{-5}$ & Conservative step size to reduce overfitting under limited supervision. \\
    Warmup steps & 5 & Short warmup relative to the total step budget. \\
    Training steps & 320 & Fixed training budget for controlled comparison. \\
    Weight decay & 0.01 & Standard L2 regularization for stability. \\
    LR scheduler & Linear & Simple decay schedule suitable for short runs. \\
    \hline
    \addlinespace[2pt]
    
    \multicolumn{3}{@{}l}{\textit{Batching}} \\
    \addlinespace[1pt]
    Per-device batch size & 2 & Constrained by GPU memory. \\
    Gradient accumulation & 4 & Effective batch size of 8. \\
    \hline
    \addlinespace[2pt]
    
    \multicolumn{3}{@{}l}{\textit{Model and objective}} \\
    Base model & LFM2.5-1.2B-Instruct & Compact instruction-tuned model enabling local experimentation~\cite{huggingfaceLiquidAILFM2512BInstructHugging}. \\
    Max sequence length & 4{,}096 & Supports long abstracts and prompt context. \\
    Response masking & Enabled & Compute loss only over assistant tokens (instruction tuning). \\
    Precision & BF16 & Efficient mixed precision on supported hardware. \\
    Training stack & Unsloth + TRL & Fast, VRAM-efficient fine-tuning with HF-compatible tooling~\cite{hfblogUnslothTRL2024,trlUnslothIntegration,unslothDocs}. \\
    \bottomrule
    \end{tabularx}
    \end{table}

%\paragraph{Supervised fine-tuning objective (response-only loss).}
%We apply response-only masking so that the loss is computed \emph{only} on assistant tokens (i.e., the label), and not on the user instruction/prompt. Concretely, we mask all tokens before the assistant marker \texttt{<|im\_start|>assistant} so that gradients update only the decision behavior rather than the prompt formatting.

\subsubsection{Dataset Partitioning}
    \begin{table}[htbp]
        \centering
        \caption{Distribution of labeled abstracts across training and test partitions.}
        \label{tab:data_splits}
        \begin{tabular}{@{}lrrrr@{}}
        \toprule
        \textbf{Dataset} & \textbf{Total} & \textbf{Exclude (0)} & \textbf{Include (1)} & \textbf{Inclusion Rate} \\
        \midrule
        Training set & 315 & 194 & 121 & 38.4\% \\
        Test set & 56 & 39 & 17 & 30.4\% \\
        \midrule
        \textbf{Total} & 371 & 233 & 138 & 37.2\% \\
        \bottomrule
        \end{tabular}
    \end{table}

The curated dataset of 371 labeled abstracts was partitioned into training and test sets, with the distribution summarized in Table~\ref{tab:data_splits}. The training set comprised 315 abstracts (194 exclusions, 121 inclusions), while the held-out test set contained 56 abstracts (39 exclusions, 17 inclusions). This split preserved approximately similar class proportions across both sets, with inclusion rates of 38.4\% and 30.4\% for training and test sets, respectively.

It is important to note that the class distribution in our curated dataset differs substantially from typical systematic review screening scenarios, where inclusion rates are typically 1--5\%. Since this was an initial exploration of feasibility and our dataset was sufficiently large, our training set was intentionally enriched with positive examples to provide sufficient signal for the model to learn inclusion criteria, addressing the challenge of extreme class imbalance in real-world screening.
\subsubsection{Training Procedure and Implementation}

%We utilized the Unsloth library, which provides optimized implementations of attention mechanisms and memory-efficient training routines, reducing memory consumption and accelerating training by approximately 2--3$\times$ compared to standard PyTorch implementations. 

Training data comprised 371 labeled abstracts, formatted as instruction-response pairs using the LFM2 chat template. Response-only masking was applied during training, computing the cross-entropy loss exclusively over assistant response tokens while masking instruction tokens. %Formally, given a training sequence $\mathbf{x} = (x_1, \ldots, x_T)$ with response tokens spanning indices $[t_r, T]$, the masked loss is: $ \mathcal{L} = -\frac{1}{T - t_r + 1} \sum_{t=t_r}^{T} \\ \log P(x_t | x_{<t}; \theta)$
This technique encourages the model to focus on learning accurate screening decisions rather than memorizing instruction templates, and has become standard practice in instruction-tuned model development~\cite{zhou2023lima}.

\subsubsection{Computational Resources}

Training was conducted on a single NVIDIA B200 GPU with 180GB memory. The 320-step training run completed in approximately 2 minutes, with peak GPU memory utilization remaining below 40GB. %The efficiency gains from LoRA and Unsloth optimization enabled rapid iteration during model development.

\subsection{Evaluation and Validation}

We employed a multi-stage evaluation strategy to assess the fine-tuned model's performance on systematic review screening, using both held-out test data and prospective screening of new abstracts.

\subsubsection{Evaluation Metrics}

Given the inherent class imbalance in systematic review screening, we selected evaluation metrics appropriate for imbalanced classification tasks. Standard accuracy can be misleading when one class dominates; therefore, we report multiple complementary metrics:

\paragraph{Balanced Accuracy.} Defined as the arithmetic mean of sensitivity (recall for the positive class) and specificity (recall for the negative class):
$\text{Balanced Accuracy} = \frac{1}{2}\left(\frac{TP}{TP + FN} + \frac{TN}{TN + FP}\right)$. This metric provides equal weight to both classes regardless of their prevalence.

\paragraph{Weighted Metrics.} We computed precision, recall, and F1-score using weighted averaging, where each class's contribution is weighted by its support (number of true instances). This approach accounts for class imbalance while still reflecting performance on the majority class.

\paragraph{Per-Class Metrics.} To provide granular insight into model behavior, we report precision, recall, and F1-score separately for both the exclude class (0) and include class (1). For systematic review screening, recall on the include class (sensitivity) is particularly critical, as missed relevant studies can compromise review validity.

\paragraph{Confusion Matrix.} We report the full confusion matrix (TP, TN, FP, FN) to enable computation of derived metrics and facilitate comparison with other screening tools.

\subsubsection{Inter--Rater Reliability}

To assess agreement between the human screener and our fine‑tuned language model, we employed several complementary measures of inter‑rater reliability.  These statistics quantify the proportion of agreement that exceeds chance expectations and help diagnose situations in which class prevalence or rater bias can unduly influence the apparent level of agreement.

\paragraph{Cohen's Kappa ($\kappa$).}  For pairwise agreement between the human decisions and each model prediction, we computed Cohen's kappa.  %Cohen introduced this chance‑corrected statistic to evaluate categorical data, adjusting the observed agreement for the amount of agreement expected under independence \cite{Cohen1960}.  Let $p_o$ be the proportion of items on which two raters agree and let $p_e$ denote the agreement expected under random ratings with the observed marginal label frequencies.  Cohen's kappa is defined as
%\begin{equation}
 %   \kappa\;=\;\frac{p_o - p_e}{1 - p_e}.
%$\end{equation}
%Following the interpretation proposed by \cite{Landis1977}, values of $\kappa$ in the range $0.61$--$0.80$ indicate ``substantial'' agreement and values above $0.80$ indicate ``almost perfect'' agreement \cite{Landis1977}.
Since $\kappa$ is sensitive to prevalence and marginal distributions \cite{Byrt1993}, we report it alongside the alternative measures described below.

\paragraph{Prevalence‑Adjusted Bias‑Adjusted Kappa (PABAK).}  One common criticism of $\kappa$ is that it decreases as the prevalence of the positive class moves away from 50\% \cite{Chen2009}.  To mitigate this sensitivity, \cite{Byrt1993} proposed a prevalence- and bias‑adjusted kappa (PABAK) that assumes equal prevalence and no bias \cite{Chen2009}. Under these assumptions and for binary classifications, PABAK reduces to $\mathrm{PABAK}\;=\;2p_o - 1.$
PABAK depends only on the observed agreement and thus provides a prevalence insensitive complement to Cohen's $\kappa$ for pairwise comparisons.

\paragraph{Gwet's Agreement Coefficient ($AC_1$).}  To further address the limitations of $\kappa$, we computed Gwet's $AC_1$ coefficient.  Gwet introduced this statistic as an alternative agreement measure whose chance‑agreement term does not depend on the marginal distributions; consequently, $AC_1$ yields more robust estimates when trait prevalence or rater classification probabilities are imbalanced \cite{Gwet2001,Gwet2002,Viswanathan2011}.  We calculated $AC_1$ for the binary ratings of the human screener and each model inference pass.  For the multi‑rater case (human plus three model passes) we computed the generalized $AC_1$ and reported $95\%$ confidence intervals obtained via a non‑parametric bootstrap over items.  %We omitted the weighted $AC_2$ statistic because our task is binary and there is no ordinal structure to the categories.

\paragraph{Fleiss' Kappa.}  To quantify agreement across more than two raters simultaneously, we computed Fleiss' kappa.  Fleiss' statistic generalizes Cohen's kappa to an arbitrary number of raters and measures how much the observed agreement exceeds what would be expected by chance across the entire rating matrix \cite{Fleiss1971}.  We evaluated Fleiss' kappa on the four‑rater matrix (human screener and three model passes) and report bootstrap‑based $95\%$ confidence intervals.

\subsubsection{Multi‑Pass Inference Protocol}

To evaluate model consistency and calibration, we performed three independent inference passes using sampling temperatures $T\in\{0.1,\,0.4,\,0.8\}$.  Lower temperatures yield near‑deterministic outputs, whereas higher temperatures introduce stochasticity that can reveal the model's uncertainty \cite{renze-2024-effect}. For each test example and temperature we recorded the model's binary prediction and computed the full suite of classification metrics against the ground‑truth human decision.  We also assessed inter‑pass reliability by computing pairwise $\kappa$ between the outputs of the different temperature settings.  

\subsubsection{Inference protocol.}  We used a structured prompt template that was recently shown to perform well in similar screening tasks in the medical domain \cite{cao2025development}. For each test example, we extracted the user prompt and ground‑truth label from the ChatML‑formatted text by locating the \texttt{user} and \texttt{assistant} markers.  Predictions were generated by applying the model's chat template to the user prompt and using greedy decoding with a short generation budget (\texttt{max\_new\_tokens}=8).  The predicted class was parsed as the first occurrence of \texttt{0} or \texttt{1} in the generated text.  To increase robustness, the evaluator also checked for decision keywords (e.g., ``include''/``exclude''); if no decision token was found, it fell back to the majority class (configured as \texttt{0}).  This procedure matches the evaluation code used to compute the agreement coefficients discussed above.

\subsection{Model Availability}
\label{sec:model_availability}

To support reproducibility, we release the fine-tuned model, dataset, and full codebase. Model: \url{https://huggingface.co/IARG-UF/lfm-2.5-1.2b-instruct_for_sys_review}. Dataset: \url{https://huggingface.co/datasets/IARG-UF/llm_sys_review}. Code: \url{https://github.com/Intelligent-Agents-Research-Group/llm_systematic_review}.

\section{Results}
   
\subsection{Baseline Performance Before Fine-Tuning}
The model's baseline performance was dismal as shown in Table \ref{tab:merged_binary_overall}. Balanced accuracy was only 53.07\%, and the per-class performance was poor (Table \ref{tab:merged_per_class}). The base model's percentage agreement with the human rater was very low (6.52\%) which is reflected in Gwet's $AC_1$ (-0.863, Table \ref{tab:agreement_three_settings}).

    \begin{table*}[htbp]
        \centering
        \caption{\textbf{Binary classification performance across evaluation settings.}
        We report results for three settings: (i) the \emph{base} model (no fine-tuning) evaluated on the full dataset, (ii) the \emph{fine-tuned} model evaluated on the held-out test split used during fine-tuning, and (iii) the \emph{fine-tuned} model evaluated on the full dataset. Metrics are percentages. \emph{Accuracy} is overall accuracy; \emph{Balanced Accuracy} is mean per-class recall (useful under strong class imbalance). We report Macro/Weighted $F_1$ and additionally Macro/Weighted $F_2$, where $F_2$ weights recall four times more than precision. Macro metrics weight classes equally; weighted metrics weight by class support.}
        \label{tab:merged_binary_overall}
        \small
        \setlength{\tabcolsep}{2.5pt}
        \renewcommand{\arraystretch}{1.15}
        \begin{tabularx}{\textwidth}{>{\raggedright\arraybackslash}X r *{6}{>{\centering\arraybackslash}c}}
        \toprule
        \textbf{Setting} & \textbf{$N$} & \textbf{Acc.} & \textbf{Bal. Acc.} & \textbf{Macro-$F_1$} & \textbf{Macro-$F_2$} & \textbf{W-$F_1$} & \textbf{W-$F_2$} \\
        \midrule
        \makecell[l]{Base (no fine-tune)\\Full dataset} &
        8{,}277 & 6.52 & 53.07 & 6.22 & 4.86 & 11.52 & 7.54 \\
        \midrule
        \makecell[l]{Fine-tuned\\Held-out test split} &
        56 & 94.64 & 94.49 & 93.77 & 94.19 & 94.68 & 94.65 \\
        \midrule
        \makecell[l]{\textbf{Fine-tuned}\\\textbf{Full dataset}} &
        8{,}277 & \textbf{86.40} & \textbf{88.78} & 48.95 & 50.41 & \textbf{92.31} & \textbf{88.48} \\
        \midrule
        \bottomrule
        \end{tabularx}
    \end{table*}
\subsection{Results with Validation Dataset}
Our fine-tuned model performed well across all metrics on the validation dataset (Table \ref{tab:merged_binary_overall}). Moreover, the per class metrics also showed strong performance for both excluded and included classifications (Table \ref{tab:merged_per_class}), and the confusion matrix showed only a 5.88\% false negatives rate (Table \ref{tab:merged_confusion_pct}). The model had strong agreement with the human coder (94.64\%) and Gwet's $AC_1$ was very good (.906, Table \ref{tab:agreement_three_settings}). 

    \begin{table*}[htbp]
        \centering
        \caption{\textbf{Per-class performance across evaluation settings.}
        We report precision, recall, and F$_1$ for each class (0=\emph{Exclude}, 1=\emph{Include}). \emph{Recall} equals \emph{per-class accuracy} (i.e., accuracy restricted to the examples of that true class): for class~0, $\mathrm{TN}/(\mathrm{TN}+\mathrm{FP})$; for class~1, $\mathrm{TP}/(\mathrm{TP}+\mathrm{FN})$. \emph{Support} is the number of true instances per class. Values are percentages.}
        \label{tab:merged_per_class}
        \small
        \setlength{\tabcolsep}{6pt}
        \renewcommand{\arraystretch}{1.10}
        \begin{tabularx}{\textwidth}{>{\raggedright\arraybackslash}X l r c c c}
        \toprule
        \textbf{Setting} & \textbf{Class} & \textbf{Support} & \textbf{Prec.} & \textbf{Rec.} & \textbf{F$_1$} \\
        \midrule
        \multirow{2}{*}{\makecell[l]{Base (no fine-tune)\\Full dataset}} 
        & 0 (Exclude) & 8{,}243 & 100.00 & 6.14 & 11.57 \\
        & 1 (Include) & 34 & 0.44 & 100.00 & 0.87 \\
        \midrule
        \multirow{2}{*}{\makecell[l]{Fine-tuned\\Held-out test split}} 
        & 0 (Exclude) & 39 & 97.37 & 94.87 & 96.10 \\
        & 1 (Include) & 17 & 88.89 & 94.12 & 91.43 \\
        \midrule
        \multirow{2}{*}{\makecell[l]{\textbf{Fine-tuned}\\\textbf{Full dataset}}} 
        & 0 (Exclude) & 8{,}243 & 99.96 & 86.38 & 92.67 \\
        & 1 (Include) & 34 & 2.69 & 91.18 & 5.22 \\
        \bottomrule
        \end{tabularx}
    \end{table*}
    
\subsection{Results Across Full Dataset}
The results from the larger corpus of all 8,277 remaining studies that were not in the training data or validation dataset showed markedly improved results compared to the base model. As shown in Table \ref{tab:merged_binary_overall}, the fine-tuned model's balanced accuracy improved by 35.71\% and the weighted F1 scores improved 80.79\%. Per-class results show that the majority class dominates aggregate performance \emph{Exclude} class, which achieves 99.96\% precision, 86.38\% recall, and 92.67\% F1 (Table~\ref{tab:merged_per_class}). In contrast, the \emph{Include} class exhibits a pronounced recall–precision trade-off: recall is high (91.18\%), indicating strong sensitivity to eligible studies, but precision is low (2.69\%), which correspondingly depresses F1 (5.22\%) (Table~\ref{tab:merged_per_class}). Since systematic review screening typically prioritizes minimizing missed inclusions over minimizing additional candidates for manual review, this emphasis on sensitivity is also reflected in the recall-weighted F2 summary reported in Table~\ref{tab:merged_binary_overall}. The confusion matrix shows that 8.82\% of the studies that should have been included were instead incorrectly classified as excluded, and only 13.62\% of cases that should have been excluded were included. Agreement with the human rater was 86.40\% and Gwet's $AC_1$ was strong (.843), even though Cohen's kappa was low (Table \ref{tab:agreement_three_settings}). Finally, we questioned if the fine-tuned model would perform well across various temperatures. The model had perfect agreement with itself across temperatures, and when the three different temperature results were analyzed in relation to the human coding, we found a strong Gwet's $AC_1$ of .842 (Table \ref{tab:agreement_multi_rater_plus_llm_consistency}).

    \begin{table}[htbp]
        \centering
        \caption{\textbf{Confusion matrices as percentages (row-normalized).}
        For each setting, we report the confusion matrix normalized by the number of true examples in each class (rows sum to 100\%). Columns correspond to predicted labels (0=\emph{Exclude}, 1=\emph{Include}); rows correspond to true labels. This presentation highlights false positive and false negative rates independently of class prevalence.}
        \label{tab:merged_confusion_pct}
        \small
        \setlength{\tabcolsep}{6pt}
        \renewcommand{\arraystretch}{1.10}
        \begin{tabular}{llcc}
            \toprule
            \textbf{Setting} & \textbf{True class} & \textbf{Pred 0} & \textbf{Pred 1} \\
            \midrule
            \multirow{2}{*}{\makecell[l]{Base (no fine-tune)\\Full dataset}}
            & True 0 (Exclude) & 6.14 & 93.86 \\
            & True 1 (Include) & 0.00 & 100.00 \\
            \midrule
            \multirow{2}{*}{\makecell[l]{Fine-tuned\\Held-out test split}} 
            & True 0 (Exclude) & 94.87 & 5.13 \\
            & True 1 (Include) & 5.88 & 94.12 \\
            \midrule
            \multirow{2}{*}{\makecell[l]{\textbf{Fine-tuned}\\\textbf{Full dataset}}} 
            & True 0 (Exclude) & 86.38 & 13.62 \\
            & True 1 (Include) & 8.82 & 91.18 \\
            \bottomrule
        \end{tabular}
    \end{table}

    \begin{table}[htbp]
        \centering
        \caption{\textbf{Model--human agreement across evaluation settings.}
        We report agreement statistics computed from the binary confusion matrix for three settings: (i) the \emph{base} model (no fine-tuning) evaluated on the full dataset, (ii) the \emph{fine-tuned} model evaluated on the held-out test split used during fine-tuning, and (iii) the \emph{fine-tuned} model evaluated on the full dataset. $P_o$ denotes observed agreement (i.e., accuracy) reported as a percentage. We additionally report Cohen's $\kappa$ (chance-corrected agreement), PABAK (prevalence-adjusted, bias-adjusted $\kappa$), and Gwet's AC1 (a chance-corrected agreement measure that is often more stable under strong prevalence imbalance).}
        \label{tab:agreement_three_settings}
        \small
        \setlength{\tabcolsep}{6pt}
        \renewcommand{\arraystretch}{1.15}
        \begin{tabularx}{\linewidth}{>{\raggedright\arraybackslash}X r c c c c}
        \toprule
        \textbf{Setting} & \textbf{$N$} & \textbf{$P_o$ (\%)} & \textbf{Cohen's $\kappa$} & \textbf{PABAK} & \textbf{Gwet AC1} \\
        \midrule
        \makecell[l]{Base (no fine-tune)\\Full dataset} & 8{,}277 & 6.52 & 0.001 & -0.870 & -0.863 \\
        \hline
        \makecell[l]{Fine-tuned\\Held-out test split} & 56 & 94.64 & 0.875 & 0.893 & 0.906 \\
        \hline
        \makecell[l]{Fine-tuned\\Full dataset} & 8{,}277 & 86.40 & 0.045 & 0.728 & 0.843 \\
        \hline
        \bottomrule
        \end{tabularx}
    \end{table}

    \begin{table}[htbp]
        \centering
        \caption{\textbf{Agreement and consistency over Phase-1 screening labels ($N=8{,}277$).}
        \emph{Multi-rater} agreement is computed across four raters: the human decision and three LLM inference runs at decoding temperatures $T\in\{0.1,0.4,0.8\}$. We report Fleiss' $\kappa$ and multi-rater Gwet's AC1 with 95\% bootstrap confidence intervals (item-wise resampling). We additionally report \emph{LLM-only} consistency as pairwise Cohen's $\kappa$ between the three temperature runs (computed over all items; no CI). Perfect pairwise agreement ($\kappa{=}1.0$) indicates identical predicted labels across temperatures in this setting.}
        \label{tab:agreement_multi_rater_plus_llm_consistency}
        \small
        \setlength{\tabcolsep}{6pt}
        \renewcommand{\arraystretch}{1.12}
        \begin{tabularx}{\linewidth}{>{\raggedright\arraybackslash}X c c c}
            \toprule
            \textbf{Statistic} & \textbf{Estimate} & \textbf{CI$_{95}$ (lo)} & \textbf{CI$_{95}$ (hi)} \\
            \midrule
            \multicolumn{4}{l}{\textbf{Human + LLM runs (4 raters)}} \\
            Fleiss' $\kappa$ & 0.603 & 0.599 & 0.607 \\
            Gwet AC1 & 0.842 & 0.834 & 0.850 \\
            \midrule
            \multicolumn{4}{l}{\textbf{LLM-only consistency (pairwise Cohen's $\kappa$ across temperatures)}} \\
            $T{=}0.1$ vs.\ $T{=}0.4$ & 1.000 & -- & -- \\
            $T{=}0.1$ vs.\ $T{=}0.8$ & 1.000 & -- & -- \\
            $T{=}0.4$ vs.\ $T{=}0.8$ & 1.000 & -- & -- \\
            \bottomrule
        \end{tabularx}
    \end{table}

\section{Discussion}
In this study we explored the extent to which fine-tuning a LLM could improve its performance and consistency in screening titles and abstracts for a specific systematic review. Our results showed that fine-tuning the model resulted in a notable performance improvements in all metrics. In the context of using LLM-based screening to save time in a systematic review screening process, we perceive the model's false positives (including studies that should be excluded) as preferable to false negatives (excluding studies that should be included), as the human rater will presumably review the false positives during full-text screening. In the context of our dataset, if we ran the model before a human screener reviewed the studies, humans would have reviewed only 13.62\% of the excluded studies, while missing 8.82\% of the studies that should have been included.  

Our results raise the question of if models can perform as well as our fine-tuned model did, how should we use them? The integration of artificial intelligence into systematic review work flows could take two forms, human in the loop or AI in the loop \cite{dellermann_hybrid_2019}. In a human in the loop scenario, AI would function as a primary screener, with a human reviewing its work. Our results show promise for this approach, but replication and more methodological work needs to be done before this can be recommended. Meanwhile, AI in the loop approaches would have the humans do the primary screening, with an AI to act as a second screener. Our results show more support for using fine-tuned LLMs for this type of approach. Having two humans screen titles and abstracts is recommended best practice from major organizations like the Campbell Collaboration \cite{aloe2024campbell} and Cochrane \cite{lefebvre2019searching}, but it is inherently time consuming and therefore expensive. Our fine-tuned LLM performed well-enough where it could be considered as a second or third screener of titles and abstracts. This assumes that at least one human has categorized all studies and the fine-tuned LLM acts as a second screener. A human could then reconcile the ratings, similar to how a human would reconcile ratings between two independent screeners. This holds the potential to save significant costs associated with two humans reviewing and categorizing a large corpus of titles and abstracts. 

\subsection{Practical Implications}
In short, our works shows the potential of embracing the \textit{context dependence} of LLMs to reduce a systematic reviewers' workload. This is consistent with prior studies that suggested that LLMs can reduce workload \cite{homiar_development_2025}. While fine-tuning an LLM may take time, it could potentially save considerable time when a systematic review contains as many studies to be reviewed as our sample did here. However, fine-tuning only saves a researcher time if the model's performance is strong enough to rationalize its use. Our model reached this level for at least considering the LLM as a second screener for our individual systematic review screening process. However, methodological work is needed to understand the generalizability of our findings, as discussed in the next section.

\subsection{Limitations and Future Directions}
Our findings highlight the need for extensive methodological work around the fine-tuning of LLMs to assist in systematic review study screening. We highlight three limitations and future directions for research: model selection, training data construction, and LLM post-training method. 

The selection of which LLM to fine-tune is not a trivial task. Since this was our initial exploration, we started with a very small LLM. We chose to use LiquidAI's LFM2.5 1.2 billion parameter model because it is very small, architecturally very fast, and can be run and fine-tuned on minimal hardware \cite{huggingfaceLiquidAILFM2512BInstructHugging}. Our results showed that even this very small LLM can perform very well when fine-tuned to do a specific task. It is an open question as to if the process would be even more effective with incrementally larger models. %In addition, our work shows that researchers need not rely on proprietary LLMs to achieve strong results in this context. This means that researchers exploring how to use LLMs in systematic review contexts can use free, open weight models, run locally on their own hardware, and publish their model weights alongside their publications as supplmentary materials to facilitate open science.

Second, extensive methodological work is needed around dataset construction, as the dataset used for training could strongly influence model performance. We were fortunate to have a corpus of more than 8500 human-coded studies from an on-going systematic review that had more than 150 studies included for full-text screening. This allowed us to construct a thorough training dataset, a unique validation dataset, and an independent full dataset on which to test the model. However, we acknowledge that our training dataset contained more included studies than a typical systematic review in education may include. This raises the question of how small of a training dataset, specifically in regards to included cases, can one use and achieve acceptable results? Moreover, we question if such a finding could generalize across models and datasets. %Research is needed to understand the impacts on model performance before researchers adopt human in the loop approach with LLMs in systematic review screening (AI is primary screener) rather than an AI in the loop approach (AI as second screener).

Finally, we used full fine-tuning, rather than an approach like low-rank adaptation (LoRA) \cite{hu2021lora} which can be more computationally efficient during training. Research is needed to better understand how other forms of LLM post-training influences LLM performance in the systematic review screening context. For example, it is possible that other approaches like direct preference optimization or various types of reinforcement learning, or even a combination of approaches into a coherent pipeline may lead to better results \cite{matsutani2025rlsqueezessftexpands,olmo20252olmo2furious}. %Research is needed to explore what works best in the context of post-training LLMs for systematic review screening workflows. 

\section{Conclusion}
Systematic reviews are time-consuming methodologies that produce the standard best-evidence often used for evidence-based decision- and policy-making. Previous research has shown mixed results as to if LLMs can facilitate the title and abstract screening process in a reliable, time-efficient matter. We sought to explore if fine-tuning a small, 1.2B parameter LLM would result in a model that not only performs well for this specific systematic review, but also performs consistently across multiple inference runs. Our results showed that fine-tuning model resulted in dramatic performance improvements and only a small false negative rate. Together, we show that there is potential for fine-tuned LLMs, trained to perform well for individual systematic reviews, to be used as at least second screeners, potentially saving time and money during the systematic review process. 

%\begin{credits}
\section*{Funding} 
This material is based upon work supported by the National Science Foundation under Grant DUE-2518159. Any opinions, findings, and conclusions or recommendations expressed in this material are those of the author(s) and do not necessarily reflect the views of the National Science Foundation.

%\end{credits}
 \bibliographystyle{unsrt}
 \bibliography{mybibliography}

@misc{huggingfaceLiquidAILFM2512BInstructHugging,
  author      = {{Liquid AI}},
  title       = {{Liquid AI}/{LFM}2.5-1.2B-Instruct $\cdot$ Hugging Face --- huggingface.co},
  howpublished= {\url{https://huggingface.co/LiquidAI/LFM2.5-1.2B-Instruct}},
  year        = {2025},
  note        = {[Accessed 19-01-2026]},
}

@article{cao2025development,
  title={Development of prompt templates for large language model--driven screening in systematic reviews},
  author={Cao, Christian and Sang, Jason and Arora, Rohit and Chen, David and Kloosterman, Robert and Cecere, Matthew and Gorla, Jaswanth and Saleh, Richard and Drennan, Ian and Teja, Bijan and others},
  journal={Annals of Internal Medicine},
  volume={178},
  number={3},
  pages={389--401},
  year={2025},
  publisher={American College of Physicians}
}

@article{borah2017analysis,
  title={Analysis of the time and workers needed to conduct systematic reviews of medical interventions using data from the PROSPERO registry},
  author={Borah, Rohit and Brown, Andrew W and Capers, Patrice L and Kaiser, Kathryn A},
  journal={BMJ open},
  volume={7},
  number={2},
  pages={e012545},
  year={2017},
  publisher={British Medical Journal Publishing Group}
}

@article{cao2025automation,
  title={Automation of Systematic Reviews with Large Language Models},
  author={Cao, Christian and Arora, Rohit and Cento, Paul and Manta, Katherine and Farahani, Elina and Cecere, Matthew and Selemon, Anabel and Sang, Jason and Gong, Ling Xi and Kloosterman, Robert and others},
  journal={medRxiv},
  pages={2025--06},
  year={2025},
  publisher={Cold Spring Harbor Laboratory Press}
}

@article{delgado2025transforming,
  title={Transforming literature screening: The emerging role of large language models in systematic reviews},
  author={Delgado-Chaves, Fernando M and Jennings, Matthew J and Atalaia, Antonio and Wolff, Justus and Horvath, Rita and Mamdouh, Zeinab M and Baumbach, Jan and Baumbach, Linda},
  journal={Proceedings of the National Academy of Sciences},
  volume={122},
  number={2},
  pages={e2411962122},
  year={2025},
  publisher={National Academy of Sciences}
}

@article{li2024evaluating,
  title={Evaluating the effectiveness of large language models in abstract screening: a comparative analysis},
  author={Li, Michael and Sun, Jianping and Tan, Xianming},
  journal={Systematic reviews},
  volume={13},
  number={1},
  pages={219},
  year={2024},
  publisher={Springer}
}

@article{Flemyng25,
  title={Position statement on artificial intelligence (AI) use in evidence synthesis across Cochrane, the Campbell Collaboration, JBI and the Collaboration for Environmental Evidence 2025},
  author={Flemyng, Ella and Noel-Storr, Anna and Macura, Biljana and Gartlehner, Gerald and Thomas, James and Meerpohl, Joerg J and Jordan, Zoe and Minx, Jan and Eisele-Metzger, Angelika and Hamel, Candyce and others},
  journal={Cochrane Database of Systematic Reviews},
  number={10},
  year={2025},
  publisher={John Wiley \& Sons, Ltd}
}

@misc{hu2021lora,
      title={LoRA: Low-Rank Adaptation of Large Language Models}, 
      author={Edward J. Hu and Yelong Shen and Phillip Wallis and Zeyuan Allen-Zhu and Yuanzhi Li and Shean Wang and Lu Wang and Weizhu Chen},
      year={2021},
      eprint={2106.09685},
      archivePrefix={arXiv},
      primaryClass={cs.CL},
      url={https://arxiv.org/abs/2106.09685}, 
}

@misc{dettmers20228bitoptimizers,
      title={8-bit Optimizers via Block-wise Quantization}, 
      author={Tim Dettmers and Mike Lewis and Sam Shleifer and Luke Zettlemoyer},
      year={2022},
      eprint={2110.02861},
      archivePrefix={arXiv},
      primaryClass={cs.LG},
      url={https://arxiv.org/abs/2110.02861}, 
}

@misc{zhou2023lima,
      title={LIMA: Less Is More for Alignment}, 
      author={Chunting Zhou and Pengfei Liu and Puxin Xu and Srini Iyer and Jiao Sun and Yuning Mao and Xuezhe Ma and Avia Efrat and Ping Yu and Lili Yu and Susan Zhang and Gargi Ghosh and Mike Lewis and Luke Zettlemoyer and Omer Levy},
      year={2023},
      eprint={2305.11206},
      archivePrefix={arXiv},
      primaryClass={cs.CL},
      url={https://arxiv.org/abs/2305.11206}, 
}

@article{dellermann_hybrid_2019,
	title = {Hybrid {Intelligence}},
	volume = {61},
	issn = {2363-7005, 1867-0202},
	url = {http://arxiv.org/abs/2105.00691},
	doi = {10.1007/s12599-019-00595-2},
	number = {5},
	urldate = {2025-08-26},
	journal = {Business \& Information Systems Engineering},
	author = {Dellermann, Dominik and Ebel, Philipp and Soellner, Matthias and Leimeister, Jan Marco},
	month = oct,
	year = {2019},
	note = {arXiv:2105.00691 [cs]},
	pages = {637--643},
}

@article{aloe2024campbell,
  title={Campbell standards: Modernizing Campbell's methodologic expectations for Campbell Collaboration intervention reviews (MECCIR)},
  author={Aloe, Ariel M and Dewidar, Omar and Hennessy, Emily A and Pigott, Terri and Stewart, Gavin and Welch, Vivian and Wilson, David B and Campbell MECCIR Working Group},
  journal={Campbell Systematic Reviews},
  volume={20},
  number={4},
  pages={e1445},
  year={2024},
  publisher={Wiley Online Library}
}

@article{lefebvre2019searching,
  title={Searching for and selecting studies},
  author={Lefebvre, Carol and Glanville, Julie and Briscoe, Simon and Littlewood, Anne and Marshall, Chris and Metzendorf, Maria-Inti and Noel-Storr, Anna and Rader, Tamara and Shokraneh, Farhad and Thomas, James and others},
  journal={Cochrane handbook for systematic reviews of interventions},
  pages={67--107},
  year={2019},
  publisher={Wiley Online Library}
}

@misc{olmo20252olmo2furious,
      title={2 OLMo 2 Furious}, 
      author={Team OLMo and Pete Walsh and Luca Soldaini and Dirk Groeneveld and Kyle Lo and Shane Arora and Akshita Bhagia and Yuling Gu and Shengyi Huang and Matt Jordan and Nathan Lambert and Dustin Schwenk and Oyvind Tafjord and Taira Anderson and David Atkinson and Faeze Brahman and Christopher Clark and Pradeep Dasigi and Nouha Dziri and Allyson Ettinger and Michal Guerquin and David Heineman and Hamish Ivison and Pang Wei Koh and Jiacheng Liu and Saumya Malik and William Merrill and Lester James V. Miranda and Jacob Morrison and Tyler Murray and Crystal Nam and Jake Poznanski and Valentina Pyatkin and Aman Rangapur and Michael Schmitz and Sam Skjonsberg and David Wadden and Christopher Wilhelm and Michael Wilson and Luke Zettlemoyer and Ali Farhadi and Noah A. Smith and Hannaneh Hajishirzi},
      year={2025},
      eprint={2501.00656},
      archivePrefix={arXiv},
      primaryClass={cs.CL},
      url={https://arxiv.org/abs/2501.00656}, 
}

@article{homiar_development_2025,
	title = {Development and evaluation of prompts for a large language model to screen titles and abstracts in a living systematic review},
	volume = {28},
	issn = {2755-9734},
	url = {https://mentalhealth.bmj.com/lookup/doi/10.1136/bmjment-2025-301762},
	doi = {10.1136/bmjment-2025-301762},
	language = {en},
	number = {1},
	urldate = {2025-12-04},
	journal = {BMJ Mental Health},
	author = {Homiar, Ava and Thomas, James and Ostinelli, Edoardo G and Kennett, Jaycee and Friedrich, Claire and Cuijpers, Pim and Harrer, Mathias and Leucht, Stefan and Miguel, Clara and Rodolico, Alessandro and Kataoka, Yuki and Takayama, Tomohiro and Yoshimura, Keisuke and So, Ryuhei and Tsujimoto, Yasushi and Yamagishi, Yosuke and Takagi, Shiro and Sakata, Masatsugu and Bašić, Đorđe and Karyotaki, Eirini and Potts, Jennifer and Salanti, Georgia and Furukawa, Toshi A and Cipriani, Andrea},
	month = jul,
	year = {2025},
	pages = {e301762},
}

@article{Chen2009,
	title = {Measuring agreement of administrative data with chart data using prevalence unadjusted and adjusted kappa},
	volume = {9},
	issn = {1471-2288},
	doi = {10.1186/1471-2288-9-5},
	language = {eng},
	journal = {BMC medical research methodology},
	author = {Chen, Guanmin and Faris, Peter and Hemmelgarn, Brenda and Walker, Robin L. and Quan, Hude},
	month = jan,
	year = {2009},
	pmid = {19159474},
	pmcid = {PMC2636838},
	pages = {5},
}

@book{Viswanathan2011,
	address = {Rockville (MD)},
	series = {{AHRQ} {Methods} for {Effective} {Health} {Care}},
	title = {Development of the {RTI} {Item} {Bank} on {Risk} of {Bias} and {Precision} of {Observational} {Studies}},
	url = {http://www.ncbi.nlm.nih.gov/books/NBK82258/},
	language = {eng},
	urldate = {2026-01-28},
	publisher = {Agency for Healthcare Research and Quality (US)},
	author = {Viswanathan, Meera and Berkman, Nancy D.},
	year = {2011},
	pmid = {22191112},
}

@article{Fleiss1971,
	title = {Measuring nominal scale agreement among many raters.},
	volume = {76},
	issn = {1939-1455, 0033-2909},
	url = {https://doi.apa.org/doi/10.1037/h0031619},
	doi = {10.1037/h0031619},
	language = {en},
	number = {5},
	urldate = {2026-01-28},
	journal = {Psychological Bulletin},
	author = {Fleiss, Joseph L.},
	month = nov,
	year = {1971},
	pages = {378--382},
}

@book{Gwet2001,
	address = {Gaithersburg, MD},
	edition = {3. ed},
	title = {Handbook of inter-rater reliability: the definitive guide to measuring the extent of agreement among raters ; [a handbook for researchers, practitioners, teachers \& students]},
	isbn = {9780970806277},
	shorttitle = {Handbook of inter-rater reliability},
	language = {eng},
	publisher = {Advanced Analytics},
	author = {Gwet, Kilem Li},
	year = {2012},
}

@inproceedings{Gwet2002,
  title={Inter-Rater Reliability: Dependency on Trait Prevalence and Marginal Homogeneity},
  author={Kilem L. Gwet},
  year={2002},
  url={https://api.semanticscholar.org/CorpusID:3621454}
}

@article{Byrt1993,
    title = {Bias, prevalence and kappa},
    journal = {Journal of Clinical Epidemiology},
    volume = {46},
    number = {5},
    pages = {423-429},
    year = {1993},
    issn = {0895-4356},
    doi = {https://doi.org/10.1016/0895-4356(93)90018-V},
    url = {https://www.sciencedirect.com/science/article/pii/089543569390018V},
    author = {Ted Byrt and Janet Bishop and John B. Carlin}
}

@misc{matsutani2025rlsqueezessftexpands,
      title={RL Squeezes, SFT Expands: A Comparative Study of Reasoning LLMs}, 
      author={Kohsei Matsutani and Shota Takashiro and Gouki Minegishi and Takeshi Kojima and Yusuke Iwasawa and Yutaka Matsuo},
      year={2025},
      eprint={2509.21128},
      archivePrefix={arXiv},
      primaryClass={cs.AI},
      url={https://arxiv.org/abs/2509.21128}, 
}

@misc{hfblogUnslothTRL2024,
  author = {Hugging Face},
  title  = {Make LLM Fine-tuning 2x faster with Unsloth and TRL},
  year   = {2024},
  howpublished = {\url{https://huggingface.co/blog/unsloth-trl}},
  note   = {Accessed: 2026-01-28}
}

@misc{hfTransformersUnslothIntegration,
  author = {Hugging Face},
  title  = {Unsloth (Transformers community integration)},
  year   = {2026},
  howpublished = {\url{https://huggingface.co/docs/transformers/community_integrations/unsloth}},
  note   = {Accessed: 2026-01-28}
}

@misc{trlUnslothIntegration,
  author = {Hugging Face},
  title  = {Unsloth Integration (TRL documentation)},
  year   = {2026},
  howpublished = {\url{https://huggingface.co/docs/trl/en/unsloth_integration}},
  note   = {Accessed: 2026-01-28}
}

@misc{unslothDocs,
  author = {Unsloth},
  title  = {Unsloth Documentation},
  year   = {2026},
  howpublished = {\url{https://unsloth.ai/docs}},
  note   = {Accessed: 2026-01-28}
}

@article{razuvayevskaya2024peftfft,
  author  = {Razuvayevskaya, Olesya and Wu, Ben and Leite, Jo{\~a}o A. and Heppell, Freddy and Srba, Ivan and Scarton, Carolina and Bontcheva, Kalina and Song, Xingyi},
  title   = {Comparison between parameter-efficient techniques and full fine-tuning: A case study on multilingual news article classification},
  journal = {PLOS ONE},
  year    = {2024},
  volume  = {19},
  number  = {5},
  pages   = {e0301738},
  doi     = {10.1371/journal.pone.0301738}
}

@misc{zhang2024,
      title={TinyLlama: An Open-Source Small Language Model}, 
      author={Peiyuan Zhang and Guangtao Zeng and Tianduo Wang and Wei Lu},
      year={2024},
      eprint={2401.02385},
      archivePrefix={arXiv},
      primaryClass={cs.CL},
      url={https://arxiv.org/abs/2401.02385}, 
}

@inproceedings{renze-2024-effect,
    title = "The Effect of Sampling Temperature on Problem Solving in Large Language Models",
    author = "Renze, Matthew",
    editor = "Al-Onaizan, Yaser  and
      Bansal, Mohit  and
      Chen, Yun-Nung",
    booktitle = "Findings of the Association for Computational Linguistics: EMNLP 2024",
    month = nov,
    year = "2024",
    address = "Miami, Florida, USA",
    publisher = "Association for Computational Linguistics",
    url = "https://aclanthology.org/2024.findings-emnlp.432/",
    doi = "10.18653/v1/2024.findings-emnlp.432",
    pages = "7346--7356"
}

@misc{belcak2025,
      title={Small Language Models are the Future of Agentic AI}, 
      author={Peter Belcak and Greg Heinrich and Shizhe Diao and Yonggan Fu and Xin Dong and Saurav Muralidharan and Yingyan Celine Lin and Pavlo Molchanov},
      year={2025},
      eprint={2506.02153},
      archivePrefix={arXiv},
      primaryClass={cs.AI},
      url={https://arxiv.org/abs/2506.02153}, 
}

@inproceedings{thakkar2024pefttradeoffs,
  author = {Thakkar, Megh and Fournier, Quentin and Riemer, Matthew D. and Chen, Pin-Yu and Zouaq, Amal and Das, Payel and Chandar, Sarath},
  title  = {A Deep Dive into the Trade-Offs of Parameter-Efficient Preference Alignment Techniques},
  booktitle = {Proceedings of the 62nd Annual Meeting of the Association for Computational Linguistics (ACL)},
  year   = {2024},
  howpublished = {\url{https://aclanthology.org/2024.acl-long.311/}}
}

@misc{lialin2024peftsurvey,
  author = {Lialin, Vladislav and Deshpande, Vihaan and Rumshisky, Anna},
  title  = {Parameter-Efficient Fine-Tuning Methods for Pretrained Language Models: A Survey},
  year   = {2024},
  eprint = {2303.15647},
  archivePrefix = {arXiv},
  primaryClass = {cs.CL}
}

\end{document}